\documentclass[lettersize,journal]{IEEEtran}
\usepackage{amsmath,amsfonts}
\usepackage{algorithmic}
\usepackage{algorithm}
\usepackage{array}
\usepackage{caption}
\usepackage[caption=false,font=normalsize,labelfont=sf,textfont=sf]{subfig}
\usepackage{textcomp}
\usepackage{stfloats}
\usepackage{url}
\usepackage{longtable}
\usepackage{diagbox}   
\usepackage{verbatim}
\usepackage{booktabs}   
\usepackage{multirow}   
\usepackage{tabularx}   
\usepackage{ragged2e}   
\usepackage{graphicx}
\usepackage{standalone}
\usepackage{forest}
\useforestlibrary{edges}
\usetikzlibrary{shadows}
\definecolor{cRoot}{RGB}{250, 220, 220}
\definecolor{cSec}{RGB}{230, 230, 250}
\definecolor{cCat}{RGB}{255, 250, 220}
\definecolor{cItem}{RGB}{225, 245, 255}
\usepackage{cite}

\newcolumntype{C}{>{\centering\arraybackslash}p{0.06\textwidth}}
\newcolumntype{Y}{>{\RaggedRight\arraybackslash}X}

\usepackage{hyperref}
\usepackage{fontawesome5}
\usepackage[table]{xcolor}
\usepackage{xstring}

\usepackage{bm}

\usepackage{amssymb}
\usepackage{tikz}
\usepackage{rotating}
\usepackage{float}
\usepackage{threeparttable}
\usepackage{placeins}
\usepackage{pifont}
\definecolor{MaxPink}{HTML}{F8D7E8}
\definecolor{MinBlue}{HTML}{DCEEFF}

\usepackage[most]{tcolorbox}
\usepackage{enumitem}
\usepackage{setspace}

\newtcolorbox{takeawaybox}[1]{
  enhanced,
  breakable,
  skin first=enhanced,
  skin middle=enhanced,
  skin last=enhanced,
  notitle after break,
  width=\linewidth,
  colback=blue!6!white,
  colframe=black!70,
  boxrule=0.8pt,
  arc=3mm,
  outer arc=3mm,
  left=4mm,
  right=4mm,
  top=4.5mm,
  bottom=-1mm,
  before skip=0.6em,
  after skip=0.3em,
  title={#1},
  fonttitle=\bfseries,
  coltitle=white,
  boxed title style={
    colback=gray!130,
    colframe=gray!80,
    boxrule=0pt,
    arc=1.5mm,
    outer arc=1.5mm,
    left=2.5mm,
    right=2.5mm,
    top=1mm,
    bottom=1mm
  },
  attach boxed title to top left={
    xshift=4mm,
    yshift=-2.2mm
  }
}

\hypersetup{
    hidelinks,
    pdftitle={ConflictVLA-Bench: Benchmarking Behavioral Responses of Vision-Language-Action Models to Premise Conflicts},
    pdfauthor={Liyu Hou, Yuan Wu, Yi Chang}
}

\begin{document}

\title{ConflictVLA-Bench: Benchmarking Behavioral Responses of Vision-Language-Action Models to Premise Conflicts}



\author{Liyu~Hou,
        Yuan~Wu, and~Yi~Chang
\thanks{Corresponding author: Yuan~Wu.}
\thanks{All authors are with the School of Artificial Intelligence, Jilin University, Changchun 130012, China (e-mail: houly25@mails.jlu.edu.cn; yuanwu@jlu.edu.cn; yichang@jlu.edu.cn). Y. Chang is also with the Engineering Research Center of Knowledge-Driven Human-Machine Intelligence, MOE, China.}
}

\markboth{Preprint}%
{Hou \MakeLowercase{\textit{et al.}}: ConflictVLA-Bench: Benchmarking Behavioral Responses of Vision-Language-Action Models to Premise Conflicts}


\maketitle
\begin{abstract}
While Vision-Language-Action (VLA) models perform strongly on manipulation tasks, their responses to invalid task premises remain underexplored. Existing evaluations of premise conflicts often focus on terminal task outcomes, yet task failure alone cannot distinguish behavioral disengagement from continued pursuit followed by an execution error. We call the latter pattern \emph{Failed Persistence}. To study this phenomenon, we introduce \textbf{ConflictVLA-Bench}, which pairs conflict rollouts with premise-consistent reference rollouts and evaluates both outcomes and execution processes. Built on LIBERO, the benchmark contains 2{,}826 prompt-conditioned conflict tasks spanning four conflict families, four structural configurations, and two prompt conditions. Across all eight VLAs, invalid premises reduce original goal completion by at least 17.3 percentage points, with the reduction reaching 56.2 percentage points for OpenVLA. Crucially, even when models succeed on premise-consistent tasks and fail on their matched conflict tasks, they often continue to approach the original targets, retain early trajectory structure, and show limited action magnitude suppression. \emph{Failed Persistence} therefore recurs across the evaluated models. Explicit premise checking does not consistently produce selective and coordinated behavioral changes. These findings show that terminal failure alone establishes neither behavioral disengagement nor refusal and that outcomes alone are insufficient for VLA evaluation. Experimental data and additional details are available on the project page:
\url{https://github.com/EmbodiedAISurvey/ConflictVLA-Bench}

\end{abstract}

\section{Introduction}
\label{sec:introduction}

\begin{figure}[t]
    \centering
    \includegraphics[
            width=\linewidth,
            trim=0mm 25mm 0mm 0mm,
            clip
        ]{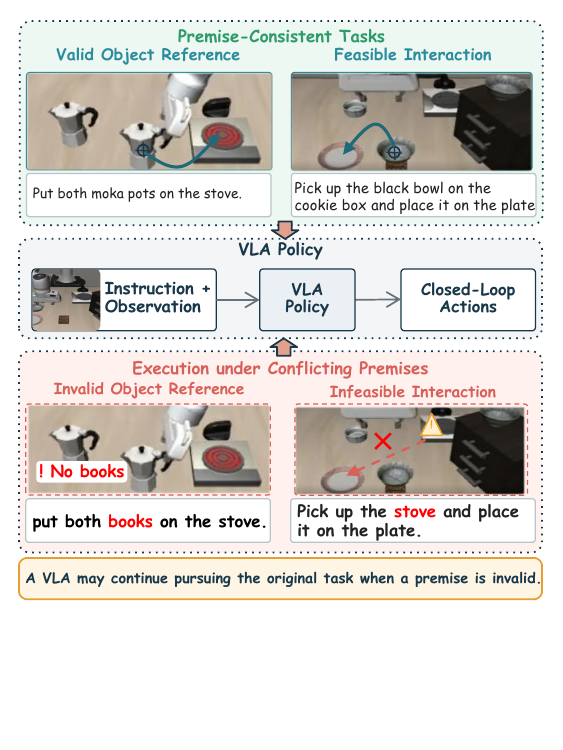}
    \caption{
    \textbf{Task competence does not guarantee reliable premise conflict response.}
A VLA may execute a task successfully under valid premises yet continue pursuing the same objective after a required premise becomes invalid.
The examples illustrate continued task-directed behavior under Object Grounding and Physical Feasibility conflicts.
    }
    \label{fig:premise_conflict_motivation}
\end{figure}

Vision-Language-Action (VLA) models integrate visual perception, language conditioning, and robot control within a single policy~\cite{zitkovich2023rt,kim2024openvla,black2024pi_0,bjorck2025gr00t}. They are a prominent paradigm for developing general purpose robotic intelligence. Established robot learning and VLA benchmarks play a central role in measuring this progress~\cite{james2020rlbench,liu2023libero,nasiriany2024robocasa,mees2022calvin,zhang2025vlaarenaopensourceframeworkbenchmarking}. Most of these benchmarks ask whether a model can complete a task whose instruction and environment provide valid premises. This setting is essential for measuring execution competence. A reliable robot, however, must not only know how to execute a task, but also determine whether the task should be carried out under current conditions~\cite{ahn2022can,huang2022inner,ren2023robots,yeke2026yes,park2023clara}.

In open world interaction, task premises do not always hold. An instruction may contain an internal contradiction, conflict with observable scene facts, or require a physically infeasible operation. As shown in Figure~\ref{fig:premise_conflict_motivation}, a policy may continue to act toward the original task even when that task no longer has a valid basis for execution. Because such actions directly affect the physical environment, continued execution can damage objects, cause unintended manipulation, or violate safety constraints~\cite{chen2026hazardarena,zhang2026safevla,fan2026safevla}. Reliable VLAs must therefore remain sensitive to the validity of task premises and adjust their subsequent behavior accordingly.

Recent studies examine VLA reliability under such conditions. LIBERO-CF evaluates language following through feasible counterfactual instructions in familiar scenes~\cite{fang2026vision}, while ICBench tests whether policies continue to execute visually supported tasks when language contradicts the scene~\cite{zhang2026restoring}. Other work trains a VLA to detect false premises and provide a clarification or correction before acting~\cite{hsieh2025teaching}. These studies provide an important foundation for evaluating premise conflict response. When a model completes the original task under a conflict, the result directly shows that the conflict does not prevent continued execution. Failure to complete the task, however, has two possible explanations. The policy may refuse to continue, or it may pursue the original objective and later fail because of perception, grounding, planning, or control errors. Thus, failure to complete the original task does not show that the policy no longer pursues it.

We call the latter pattern \textbf{\emph{Failed Persistence}}. In a conflict task, the original task remains incomplete, yet the policy continues to exhibit behavior directed toward the invalid objective. For example, a model may approach the same target and follow a similar early trajectory as in normal execution, but later fail during grasping or control. Determining whether a conflict failure exhibits Failed Persistence requires process evidence, including behavior directed toward the target, trajectory retention, and action suppression.

\begin{figure*}[t]
    \centering
    \includegraphics[
            width=\textwidth,
            trim=13mm 7mm 2mm 2mm,
            clip
        ]{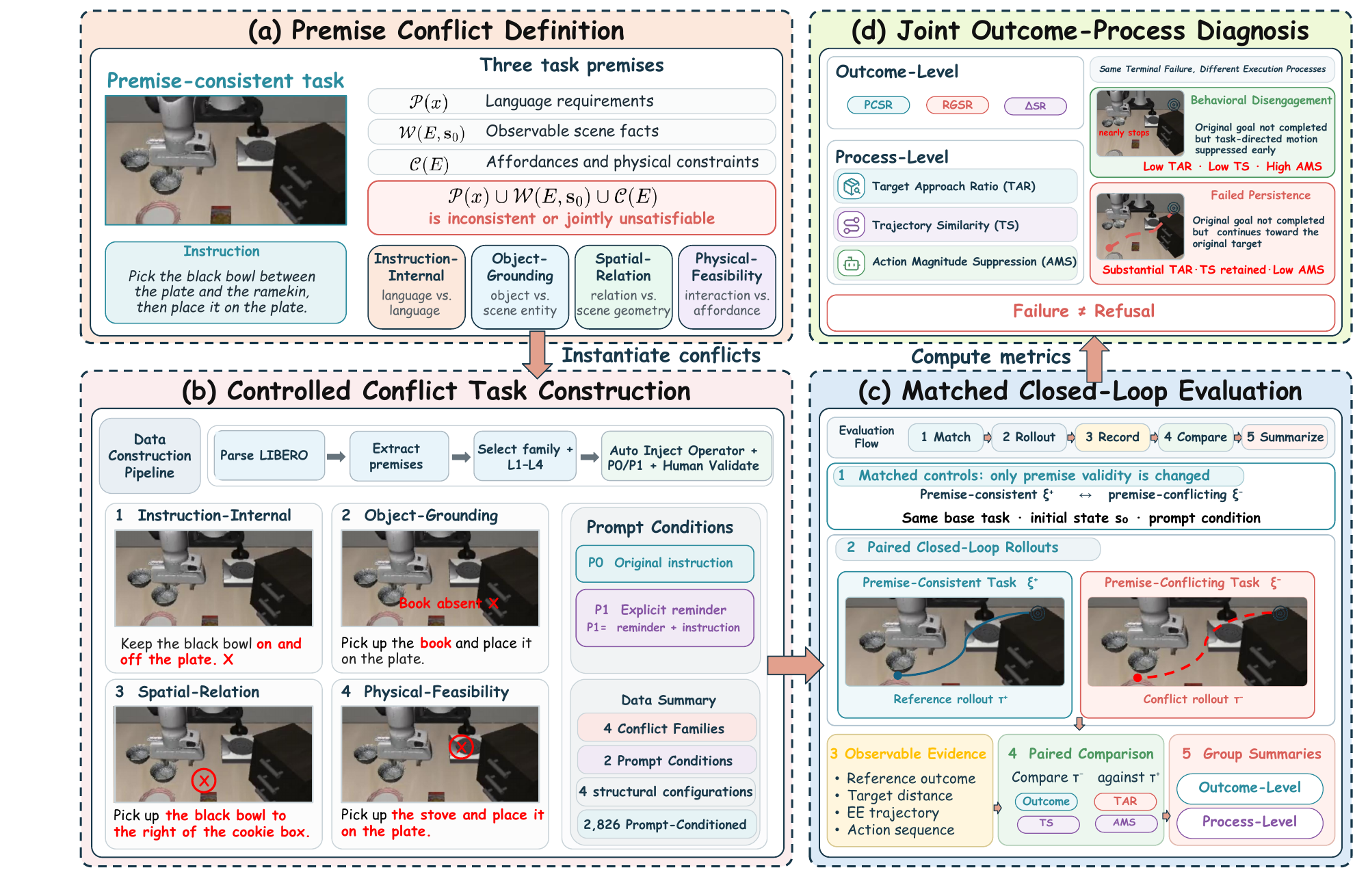}
    \caption{
    \textbf{Overview of ConflictVLA-Bench.}
(a) We define four premise conflict families.
(b) Controlled conflict tasks are constructed from LIBERO across four structural configurations and two prompt conditions.
(c) Each conflict rollout is compared with a matched premise-consistent reference rollout.
(d) Outcome and process evidence jointly diagnose behavioral disengagement and Failed Persistence.
    }
    \label{fig:conflictvla_overview}
\end{figure*}

To study this behavioral pattern, we introduce \textbf{ConflictVLA-Bench}, a suite for evaluating VLA behavior during closed-loop execution after task premises fail. As illustrated in Figure~\ref{fig:conflictvla_overview}, each conflict rollout is paired with a reference rollout produced by the same model for the same base task, initial state, and prompt condition under valid premises. The reference rollout records how the model executes the task under normal conditions, whereas the conflict rollout records its behavior after a required premise fails. When diagnosing conflict failures, we use only pairs in which the reference rollout succeeds, which confirms that the model can complete the base task. Comparing the paired rollouts allows us to examine not only whether the original goal is completed, but also whether the model approaches the same target, retains a similar trajectory, or suppresses its action magnitude. The protocol evaluates observable behavior and does not infer whether the model internally recognizes or understands the conflict.

ConflictVLA-Bench builds on LIBERO~\cite{liu2023libero} and generates conflict tasks by injecting controlled inconsistencies, invalid references, and unsatisfied premises. It covers four representative conflict families: Instruction Internal Conflict, Object Grounding Conflict, Spatial Relation Conflict, and Physical Feasibility Conflict. Each family contains four structural configurations, from L1 to L4, and is evaluated under two prompt conditions, with or without an explicit instruction to check task premises. The benchmark contains 2{,}826 conflict tasks conditioned on these prompts.

We evaluate eight representative VLAs and find that the reduction in reference goal completion under conflicts does not consistently correspond to behavioral disengagement. Even when the original task fails, models often continue to approach the original target, retain high similarity to the reference trajectory during early execution, and show limited action suppression. Among matched pairs in which the premise-consistent rollout succeeds and the conflict rollout fails, the mean Early20 trajectory similarity, computed over the first 20 executed actions, exceeds 0.650 for every model. Failed Persistence therefore recurs across models, although it does not describe every failure. Explicit instructions to check task premises also fail to consistently reduce retention of the original task behavior and, for some models, coincide with poorer performance on tasks with valid premises. Overall, normal task competence does not guarantee that a VLA stops pursuing an objective after its premises fail.

Our main contributions are as follows:

\begin{itemize}

\item We identify and characterize a critical behavioral failure pattern under premise conflicts. A policy may fail to complete the original task while continuing to exhibit behavior directed toward its invalid objective. We call this pattern \textbf{\emph{Failed Persistence}} and show why terminal failure alone cannot establish refusal.

\item We introduce \textbf{ConflictVLA-Bench}, which contains 2{,}826 conflict tasks across four conflict families, four structural configurations, and two prompt conditions. We also establish a paired evaluation protocol with capability qualification that combines outcome and process evidence from tasks with valid and invalid premises.

\item We systematically evaluate eight representative VLAs. Failed Persistence recurs across the evaluated models, revealing limited premise conflict response, while explicit instructions to check task premises do not consistently produce selective and coordinated behavioral changes.

\end{itemize}

\section{Related Work}
\label{sec:related_work}

\subsection{VLA Benchmarks and Diagnostic Evaluation}

Benchmarks such as RLBench~\cite{james2020rlbench}, LIBERO~\cite{liu2023libero}, RoboCasa~\cite{nasiriany2024robocasa}, and CALVIN~\cite{mees2022calvin} establish reproducible settings for language conditioned manipulation and measure policy competence when instructions and scenes specify executable goals. Building on this foundation, LIBERO-PRO~\cite{zhou2025liberoprorobustfairevaluation} and LIBERO-Plus~\cite{fei2025liberoplusindepthrobustnessanalysis} evaluate robustness under controlled variations, while VLA-Arena~\cite{zhang2025vlaarenaopensourceframeworkbenchmarking} studies structured task difficulty together with language and visual perturbations. VLA-Trace~\cite{shi2026vla} connects representation analysis to rollout probes for diagnosing grounding and shortcut dependence, while safety evaluations examine semantic risks~\cite{chen2026hazardarena} or trajectory violations that can remain hidden by successful task completion~\cite{fan2026safevla}. These efforts address complementary capability boundaries in competence, robustness, diagnosis, and safety. ConflictVLA-Bench focuses on a complementary condition in which a premise required by the reference task no longer holds. We ask whether the policy still exhibits reference task behavior after that premise becomes invalid, using matched closed loop evidence rather than interpreting task success or failure in isolation.

\subsection{Premise Evaluation and Conflict Response}

Research on embodied agents has studied affordance grounded planning~\cite{ahn2022can}, environmental feedback~\cite{huang2022inner}, help seeking~\cite{ren2023robots}, and abstention~\cite{yeke2026yes} when a request cannot be resolved; PCBench~\cite{li2025don} evaluates premise critique through explicit textual judgments from language models. In VLA manipulation, LIBERO-CF~\cite{fang2026vision} tests language following with feasible counterfactual instructions in familiar scenes; ICBench~\cite{zhang2026restoring} probes language action coupling with controlled contradictions and supplements task outcomes with representative trajectories; and Hsieh et al.~\cite{hsieh2025teaching} train an explicit interface that detects false premises and produces clarifications or corrections before acting. These works establish useful settings under their respective task and response interfaces. We study a complementary setting in which a general VLA is evaluated through closed loop actions without requiring a standardized refusal output. Completing the reference goal under a conflict directly shows continued execution, but failure alone does not show that the policy stopped pursuing the task. We therefore match each conflict rollout with its premise-consistent counterpart, restrict failure diagnosis to cases in which that counterpart succeeds, and jointly examine terminal outcomes, target approach, trajectory retention, and action suppression. This protocol characterizes \emph{Failed Persistence} as an observable behavior pattern; it does not establish latent conflict understanding, language refusal, or general safety.

\section{ConflictVLA Benchmark}

\subsection{Task Definition}

We formalize premise conflicts and matched closed-loop rollouts to support benchmark construction and behavioral analysis. Let \(\pi_\theta\) denote a VLA policy with parameters \(\theta\). A task instance is represented as \(\xi=(x,E,\mathbf{s}_0)\), where \(x\) is the language instruction, \(E\) is the scene specification, and \(\mathbf{s}_0\) is the initial state. Let \(H\) denote the number of executed action steps in a rollout. During closed-loop execution, at each step \(t\in\{0,\ldots,H-1\}\), the policy produces an action \(\mathbf{a}_t\) from \(x\) and the available visual and proprioceptive interaction history. Together with the environment dynamics, these actions induce the rollout
\[
\tau(\xi)
=
(\mathbf{s}_0,\mathbf{o}_0,\mathbf{a}_0,
\mathbf{s}_1,\mathbf{o}_1,\ldots,
\mathbf{a}_{H-1},\mathbf{s}_H,\mathbf{o}_H),
\]
where \(\mathbf{s}_t\), \(\mathbf{o}_t\), and \(\mathbf{a}_t\) denote the environment state, policy observation, and executed action at step \(t\), respectively.

Let \(\mathcal{P}(x)\), \(\mathcal{W}(E,\mathbf{s}_0)\), and \(\mathcal{C}(E)\) denote sets of temporally grounded logical conditions over a candidate execution, encoding the task premises and requirements, the relevant initial scene facts, and the affordance and physical constraints, respectively.
We call \(\xi\) premise-conflicting, denoted by \(\xi^{-}\), when no feasible execution from \(\mathbf{s}_0\) in \(E\) can jointly satisfy all conditions in \(\mathcal{P}(x)\cup\mathcal{W}(E,\mathbf{s}_0)\cup\mathcal{C}(E)\).
Otherwise, the task is premise-consistent and is denoted by \(\xi^{+}\).

For each conflicting task \(\xi^{-}\), we construct a matched premise-consistent counterpart \(\xi^{+}\) from the same underlying task, initial state, and prompt condition. Executing \(\pi_\theta\) on the two tasks produces the matched rollouts \(\tau^{-}=\tau(\xi^{-})\) and \(\tau^{+}=\tau(\xi^{+})\). The premise-consistent counterpart is used only by the evaluator and is not provided to the policy during execution of \(\xi^{-}\).

Let \(D=\{(\tau_i^{+},\tau_i^{-})\}_{i=1}^{N}\) denote an evaluation set of \(N\) matched rollout pairs. For pair \(i\), let \(y_i^{+}\in\{0,1\}\) indicate whether the premise-consistent rollout completes its task and let \(y_i^{-}\in\{0,1\}\) indicate whether the conflict rollout completes the original goal. We define
\begin{equation}
\begin{aligned}
D_{\mathrm{Q}}
&=
\{(\tau_i^{+},\tau_i^{-})\in D \mid y_i^{+}=1\},\\
D_{\mathrm{QF}}
&=
\{(\tau_i^{+},\tau_i^{-})\in D_{\mathrm{Q}}
\mid y_i^{-}=0\},
\end{aligned}
\label{eq:qualified_sets}
\end{equation}
where \(D_{\mathrm{Q}}\) is the capability-qualified set and \(D_{\mathrm{QF}}\) is its qualified failure subset. Membership in \(D_{\mathrm{QF}}\) confirms successful premise-consistent execution for the matched case and failure to complete the original goal under conflict, but it does not by itself establish observable behavioral disengagement. We therefore examine the associated process measures defined below. More generally, observable evidence of refusal requires both demonstrated premise-consistent capability and coordinated suppression of behavior associated with the original task.

\subsection{Task Design}

We construct ConflictVLA-Bench by extending
LIBERO~\cite{liu2023libero}, whose standardized BDDL specifications and
controllable simulator configurations support controlled conflict
construction and reproducible trajectory evaluation. Starting from an
executable LIBERO task, we construct a conflict task by invalidating one
or more premises required for execution while preserving the initial scene
configuration and retaining the original task as its matched
premise-consistent counterpart. The shared LIBERO environment and task
protocol enable comparisons across model families.

\subsubsection{Conflict Suite}

We define four representative premise conflict families. Each instance is
assigned according to its designated construction operator and primary
source of premise invalidity. The families target different sources of
invalidity and the resulting observable policy responses.

\textbf{Instruction-Internal Conflict.}
We introduce mutually incompatible goals, actions, states, or constraints
into an instruction, creating requirements that cannot be satisfied
simultaneously. All evidence required to establish the conflict is
contained in the language instruction, so this family does not require
consistency checking between language and visual observations. This family
evaluates whether execution toward the instructed task is suppressed when
the instruction is internally inconsistent.

\textbf{Object-Grounding Conflict.}
Object grounding conflicts are formed by replacing the manipulated object,
destination object, or receptacle with referents that are absent from or
cannot be validly grounded in the current scene. This family examines
whether models align linguistic references with perceived entities rather
than redirecting an invalid reference to a familiar or canonical object,
and whether they suppress execution when no valid referent is available.

\textbf{Spatial-Relation Conflict.}
We alter a spatial premise, landmark, or relational referring expression
so that a relation presupposed by the instruction is false in the initial
scene. The conflict concerns a relation that must hold before execution,
rather than a spatial goal that the requested action is intended to
establish. The referenced entities remain present and individually
groundable, so the invalidity arises from the relation among them. This
distinguishes spatial relation conflicts from object grounding conflicts.
This family evaluates whether policy behavior is conditioned on
directional, relational, and landmark evidence.

\textbf{Physical-Feasibility Conflict.}
We modify the manipulated object, destination, or requested action such
that the resulting task violates object affordances or manipulation
constraints. The referenced entities remain present and groundable, but
the requested interaction cannot be physically realized under the
constraints of the environment. This family evaluates whether execution
is suppressed for infeasible combinations of objects and actions or
objects and destinations.

\subsubsection{Structural Configurations}

We vary two structural factors in conflict construction, namely the number
of manipulation steps and the number of conflict factors. The four
resulting configurations are instantiated using operators designed for
each conflict family.

\textbf{Level 1 (L1) Single-Step Atomic Conflict:}
L1 combines (i)~one manipulation step and (ii)~one localized conflict
factor. The conflict can be identified without depending on additional
subgoals.

\textbf{Level 2 (L2) Single-Step Compound Conflict:}
L2 combines (i)~one manipulation step and (ii)~multiple conflict factors
that jointly affect the same manipulation requirement.

\textbf{Level 3 (L3) Multi-Step Atomic Conflict:}
L3 combines (i)~multiple ordered manipulation steps and (ii)~one conflict
localized to a particular subgoal. The remaining task components preserve
a plausible execution sequence.

\textbf{Level 4 (L4) Multi-Step Compound Conflict:}
L4 combines (i)~multiple ordered manipulation steps and (ii)~multiple
interacting conflict factors distributed across different task components.
The conflict evidence and task dependencies may therefore span a longer execution sequence. These configurations describe construction structure rather than an assumed empirical difficulty ordering.

\subsubsection{Prompt Conditions}

For each conflict task, we define two prompt conditions to compare
observable behavior with and without explicit guidance to check task
premises. Under the standard prompt condition (P0), the model receives
the conflicting instruction in the standard VLA task format. It is not
informed that the task may contain a premise conflict, nor is it prompted
to suppress execution. P0 serves as the primary evaluation condition and
measures the model's response without prompt intervention.

Under the explicit premise checking condition (P1), we prepend a prompt
that asks the model to check the validity of the task premises and suppress
execution if a conflict is found. We apply both prompt conditions to the
matched premise-consistent tasks to control for changes caused by the
prompt. Comparing P0 and P1 assesses whether explicit guidance selectively
changes behavior under conflict while preserving premise-consistent task
execution.

\subsubsection{Construction and Validation Pipeline}

The construction pipeline comprises five main steps. First, we parse the
original LIBERO instruction, BDDL goal specification, and scene
configuration. Second, we extract task entities, goal predicates, and
relevant spatial and affordance constraints. Third, we select a conflict
operator according to the designated conflict family and structural
configuration and inject the corresponding conflict while preserving the
underlying scene configuration. Fourth, we assign paired identifiers, link
each generated conflict case to its premise-consistent counterpart, and
instantiate it under the P0 and P1 conditions.

Finally, automatic checks are followed by manual validation in two stages.
We first inspect every base conflict case to confirm that the intended
conflict is valid and observable, its conflict family and structural
configuration are correct, and its matched counterpart remains valid. We
then recheck all flagged cases and a stratified sample covering every
conflict family and structural configuration. Cases are removed when the
conflict is invalid or cannot be verified from the observable task context,
or when the corresponding premise-consistent counterpart is no longer
valid. Cases with correctable annotation errors are revised and
revalidated. For example, if an Object Grounding Conflict is designed
around a missing object but that object is actually present in the scene,
the case does not constitute a valid conflict and is therefore removed.
This process retains \(1{,}413\) base conflict cases, yielding \(2{,}826\)
prompt-conditioned instances under P0 and P1.
\subsection{Evaluation Metrics}

Using the evaluation set \(D\) and the outcome indicators \(y_i^{+}\) and
\(y_i^{-}\) defined above, we report complementary outcome and process
measures.

\subsubsection{Outcome Measures}

The premise-consistent success rate (PCSR), reference goal success rate
(RGSR), and their difference are
\(\mathrm{PCSR}(D)=\frac{1}{N}\sum_{i=1}^{N}y_i^{+}\),
\(\mathrm{RGSR}(D)=\frac{1}{N}\sum_{i=1}^{N}y_i^{-}\), and
\(\Delta\mathrm{SR}(D)=\mathrm{PCSR}(D)-\mathrm{RGSR}(D)\).
Here, RGSR measures whether the conflict rollout completes the original
task goal used by the evaluator. Because these rates are computed over
\(D\), a premise-consistent rollout reused by multiple conflict instances
contributes once to each matched pair. For
\(D_{\mathrm{Q}}\neq\varnothing\), the failure rate among
capability-qualified cases is
\(\mathrm{FR}(D)=\frac{|D_{\mathrm{QF}}|}{|D_{\mathrm{Q}}|}\).

\subsubsection{Process Measures}

We use \(\epsilon=10^{-8}\) for numerical stability and define
\(\operatorname{clip}_{[0,1]}(z)=\min(1,\max(0,z))\).
For matched pair \(i\) and \(r\in\{+,-\}\), let \(H_i^r\) denote the
number of executed actions in rollout \(\tau_i^r\), where \(+\) and \(-\)
index the premise-consistent and conflict rollouts, respectively.

\textbf{Target Approach Ratio (TAR).}
Let \(\mathcal{Q}_i\) denote the set of objects manipulated in the original
task, \(\mathbf{p}_{i,t}^{r}\) the end effector position, and
\(\mathbf{q}_{i,t,k}^{r}\) the position of object
\(k\in\mathcal{Q}_i\) at recorded state \(t\). We define
\begin{equation}
\begin{gathered}
d_{i,t}^{r}
=
\min_{k\in\mathcal{Q}_i}
\left\lVert
\mathbf{p}_{i,t}^{r}-\mathbf{q}_{i,t,k}^{r}
\right\rVert_2,\\
\mathcal{A}_i^{r}
=
\max\left\{
d_{i,0}^{r}
-
\min_{0\leq t\leq H_i^{r}}d_{i,t}^{r},
0
\right\},\\
\mathrm{TAR}_i
=
\operatorname{clip}_{[0,1]}
\left(
\frac{\mathcal{A}_i^{-}}
{\max(\mathcal{A}_i^{+},\epsilon)}
\right).
\end{gathered}
\label{eq:tar}
\end{equation}
TAR is computed only when the two rollouts have identical target sets
defined by the evaluator and computable target distances.

\textbf{Trajectory Similarity (TS).}
Let
\(\mathbf{P}_i^{r}
=(\mathbf{p}_{i,0}^{r},\ldots,\mathbf{p}_{i,H_i^{r}}^{r})\)
denote the end effector trajectory. Using dynamic time warping with
Euclidean local costs, let
\(\mathcal{D}_{\mathrm{DTW}}(\mathbf{P}_i^{-},\mathbf{P}_i^{+})\)
denote the accumulated distance and \(L_i\) the corresponding alignment
path length for the trajectory segment being evaluated. We use
\(\sigma=0.10\,\mathrm{m}\) and compute
\begin{equation}
\mathrm{TS}_i
=
\exp\left(
-\frac{
\mathcal{D}_{\mathrm{DTW}}
(\mathbf{P}_i^{-},\mathbf{P}_i^{+})
}{
\sigma L_i
}
\right).
\label{eq:ts}
\end{equation}
TS uses the complete trajectories. Early20 TS uses the initial position
and the positions produced by the first
\(\min(20,H_i^{r})\) executed actions of each rollout.

\textbf{Action Magnitude Suppression (AMS).}
We use the first six continuous motion dimensions of the common LIBERO
action representation and exclude the gripper command. Let
\(a_{i,t,j}^{r}\) denote action dimension
\(j\in\{1,\ldots,6\}\) at step \(t\) of \(\tau_i^{r}\). We compute the
RMS scale of each dimension from the premise-consistent rollout as
\[
\gamma_{i,j}
=
\sqrt{
\max\left(
\frac{1}{H_i^{+}}
\sum_{t=0}^{H_i^{+}-1}
\left(a_{i,t,j}^{+}\right)^2,
\epsilon
\right)
}.
\]
The normalized action intensity is
\[
\mathcal{I}(\tau_i^{r})
=
\frac{1}{H_i^{r}}
\sum_{t=0}^{H_i^{r}-1}
\left\lVert
\left(
\frac{a_{i,t,1}^{r}}{\gamma_{i,1}},
\ldots,
\frac{a_{i,t,6}^{r}}{\gamma_{i,6}}
\right)
\right\rVert_2.
\]
For pairs with valid, nonempty action sequences and
\(\mathcal{I}(\tau_i^{+})>\epsilon\), we define
\begin{equation}
\mathrm{AMS}_i
=
\operatorname{clip}_{[0,1]}
\left(
1-
\frac{\mathcal{I}(\tau_i^{-})}
{\mathcal{I}(\tau_i^{+})}
\right).
\label{eq:ams}
\end{equation}
A higher AMS indicates a stronger reduction in action magnitude relative
to the matched premise-consistent rollout. TAR, TS, and AMS characterize
complementary observable behaviors and are interpreted jointly rather
than as individual evidence of behavioral disengagement.

\section{Experiment}
\label{sec:results}

\subsection{Experimental Setup}

We evaluate VLAs on ConflictVLA-Bench. To systematically assess the premise-conflict response capability of current VLAs, we evaluate 8 representative architectures spanning 4 families: (1) \emph{Tokenized Autoregressive VLAs}: OpenVLA~\cite{kim2024openvla} and $\pi_0$-FAST~\cite{pertsch2025fast}; (2) \emph{Continuous-Action VLAs}: OpenVLA-OFT~\cite{kim2025fine}, $\pi_0$~\cite{black2024pi_0}, and $\pi_{0.5}$~\cite{intelligence2025pi_}; (3) \emph{World Model-based VLAs}: VLA-JEPA~\cite{sun2026vla} and UniVLA~\cite{bu2025univla}; and (4) \emph{Generalist Robot Foundation Models}: GR00T N1.7~\cite{bjorck2025gr00t}. 

For each model, we use the released LIBERO-compatible checkpoint and follow its native inference protocol. We evaluate each of the 2{,}826 prompt-conditioned conflict tasks on five initial states and each of the 40 premise-consistent tasks on the same five states under both P0 and P1. This yields 14{,}130 conflict rollouts and 400 premise-consistent rollouts, for a total of 14{,}530 rollouts per model. Each conflict rollout is paired with the premise-consistent rollout from the same base task, initial state, and prompt condition. A premise-consistent rollout may be reused for multiple conflict tasks derived from the same base task, but every matched pair is evaluated separately. All metrics are computed at the episode level before aggregation, with pairwise metrics evaluated on these exact matches. Unless otherwise specified, the main tables report results under the P0 prompt condition, while P1 is analyzed separately in RQ4.

\begin{table}[t]
\centering
\caption{
Overall outcome and process measures (\%) for eight VLAs.
Values that are \underline{underlined} are the most favorable for each metric.
}
\label{tab:overall_conflict_metrics}

\fontsize{6.4}{7.2}\selectfont
\setlength{\tabcolsep}{1.2pt}
\renewcommand{\arraystretch}{1.03}

\resizebox{\columnwidth}{!}{%
\begin{tabular}{lccc@{\hspace{8pt}}cccc}
\toprule
\multirow{2}{*}{Model}
& \multicolumn{3}{c}{Outcome}
& \multicolumn{4}{c}{Process} \\
\cmidrule(lr){2-4}\cmidrule(lr){5-8}
& PCSR & RGSR & $\Delta$SR
& TAR & Early20 TS & TS & AMS \\
\midrule
OpenVLA & 78.7 & \underline{22.5} & \underline{56.2} & \underline{76.5} & \underline{78.4} & \underline{47.1} & \underline{27.6} \\
OpenVLA-OFT & 98.3 & 71.1 & 27.1 & 89.4 & 88.6 & 74.4 & 6.7 \\
\(\pi_0\) & 93.1 & 57.6 & 35.5 & 86.9 & 83.8 & 59.0 & 8.0 \\
\(\pi_0\)-FAST & 83.4 & 44.8 & 38.6 & 85.2 & 88.6 & 57.6 & 10.2 \\
\(\pi_{0.5}\) & 94.5 & 77.2 & 17.3 & 89.1 & 85.9 & 71.4 & 6.9 \\
GR00T N1.7 & 92.1 & 63.2 & 29.0 & 86.9 & 86.6 & 63.0 & 10.1 \\
VLA-JEPA & \underline{99.0} & 69.1 & 29.9 & 89.6 & 87.6 & 71.4 & 13.2 \\
UniVLA & 84.1 & 59.5 & 24.6 & 88.5 & 88.6 & 69.1 & 9.5 \\
\bottomrule
\end{tabular}%
}

\renewcommand{\arraystretch}{1.0}
\end{table}

\begin{table*}[t]
\centering
\caption{
Outcome and process measures of eight VLAs across four premise-conflict families.
\protect\colorbox{MaxPink}{Pink} marks the maximum within each metric column, and
\protect\colorbox{MinBlue}{blue} marks the minimum.
All metrics are reported in \%.
}
\label{tab:model_by_conflict_taxonomy_main}
\scriptsize
\setlength{\tabcolsep}{2pt}
\renewcommand{\arraystretch}{1.15}
\resizebox{\textwidth}{!}{%
\begin{tabular}{l*{24}{r}}
\toprule
\multirow{2}{*}{\textbf{Model}}
& \multicolumn{6}{c}{\textbf{Instruction}}
& \multicolumn{6}{c}{\textbf{Grounding}}
& \multicolumn{6}{c}{\textbf{Spatial}}
& \multicolumn{6}{c}{\textbf{Physical}} \\
\cmidrule(lr){2-7}
\cmidrule(lr){8-13}
\cmidrule(lr){14-19}
\cmidrule(lr){20-25}
& PCSR & RGSR & $\Delta$SR & TAR & TS & AMS
& PCSR & RGSR & $\Delta$SR & TAR & TS & AMS
& PCSR & RGSR & $\Delta$SR & TAR & TS & AMS
& PCSR & RGSR & $\Delta$SR & TAR & TS & AMS \\
\midrule
OpenVLA
& \cellcolor{MinBlue}{78.2} & \cellcolor{MinBlue}{29.0} & \cellcolor{MaxPink}{49.2} & \cellcolor{MinBlue}{83.6} & \cellcolor{MinBlue}{49.9} & \cellcolor{MaxPink}{29.2}
& 81.8 & \cellcolor{MinBlue}{32.0} & \cellcolor{MaxPink}{49.8} & \cellcolor{MinBlue}{88.1} & \cellcolor{MinBlue}{53.7} & \cellcolor{MaxPink}{27.5}
& \cellcolor{MinBlue}{73.7} & \cellcolor{MinBlue}{12.1} & \cellcolor{MaxPink}{61.6} & \cellcolor{MinBlue}{60.7} & \cellcolor{MinBlue}{44.3} & \cellcolor{MaxPink}{26.6}
& \cellcolor{MinBlue}{81.0} & \cellcolor{MinBlue}{16.5} & 64.5 & 73.1 & \cellcolor{MinBlue}{40.2} & \cellcolor{MaxPink}{26.9} \\

OpenVLA-OFT
& 97.5 & 84.5 & 13.0 & \cellcolor{MaxPink}{97.5} & \cellcolor{MaxPink}{84.8} & 5.6
& 98.0 & \cellcolor{MaxPink}{79.0} & 19.0 & 94.5 & \cellcolor{MaxPink}{78.8} & \cellcolor{MinBlue}{4.8}
& \cellcolor{MaxPink}{100.0} & 53.0 & 47.0 & 80.1 & 63.7 & 10.6
& 97.5 & \cellcolor{MaxPink}{68.0} & \cellcolor{MinBlue}{29.5} & \cellcolor{MaxPink}{86.3} & \cellcolor{MaxPink}{70.3} & \cellcolor{MinBlue}{5.8} \\

$\pi_0$
& 92.1 & 74.5 & 17.6 & 95.1 & 68.5 & 6.9
& 92.0 & 66.0 & 26.0 & 93.9 & 65.8 & 7.4
& 96.0 & 40.1 & 55.9 & 78.1 & 48.7 & 9.3
& 92.5 & 49.4 & 43.1 & 81.5 & 52.6 & 8.7 \\

$\pi_0$-FAST
& 86.8 & 45.5 & 41.2 & 88.9 & 59.9 & 10.3
& \cellcolor{MinBlue}{77.3} & 45.2 & 32.1 & 90.7 & 57.5 & 10.6
& 84.0 & 68.6 & 15.5 & 89.5 & \cellcolor{MaxPink}{72.4} & \cellcolor{MinBlue}{6.3}
& 85.5 & 20.8 & \cellcolor{MaxPink}{64.8} & \cellcolor{MinBlue}{71.6} & 41.0 & 13.4 \\

$\pi_{0.5}$
& 95.0 & \cellcolor{MaxPink}{93.8} & \cellcolor{MinBlue}{1.2} & 97.2 & 82.3 & \cellcolor{MinBlue}{4.4}
& 96.5 & 76.8 & 19.8 & 91.6 & 74.6 & 6.6
& 89.9 & \cellcolor{MaxPink}{79.4} & \cellcolor{MinBlue}{10.6} & \cellcolor{MaxPink}{90.0} & 67.7 & 7.0
& 96.2 & 59.0 & 37.3 & 77.2 & 60.6 & 9.6 \\

GR00T N1.7
& 90.2 & 78.8 & 11.5 & 95.0 & 72.7 & 8.1
& 93.8 & 78.5 & \cellcolor{MinBlue}{15.2} & 94.0 & 74.0 & 8.2
& 90.2 & 44.8 & 45.4 & 76.9 & 48.6 & 12.2
& 94.2 & 50.0 & 44.2 & 81.5 & 56.3 & 12.1 \\

VLA-JEPA
& \cellcolor{MaxPink}{98.4} & 81.9 & 16.5 & 96.7 & 79.6 & 11.8
& \cellcolor{MaxPink}{99.2} & 76.4 & 22.8 & \cellcolor{MaxPink}{94.7} & 78.7 & 10.2
& \cellcolor{MaxPink}{100.0} & 57.9 & 42.1 & 82.8 & 62.6 & 14.4
& \cellcolor{MaxPink}{98.9} & 55.2 & 43.7 & 81.0 & 60.5 & 16.8 \\

UniVLA
& 86.0 & 71.5 & 14.5 & 95.9 & 76.5 & 8.0
& 79.8 & 60.2 & 19.5 & 91.9 & 68.4 & 9.5
& 83.5 & 52.2 & 31.2 & 83.9 & 69.8 & 12.4
& 87.0 & 54.0 & 33.0 & 82.0 & 61.6 & 8.1 \\
\bottomrule
\end{tabular}%
}
\renewcommand{\arraystretch}{1.0}
\end{table*}

\subsection{Overall Results}

Across eight representative VLAs, invalid premises expose a consistent mismatch between terminal outcome sensitivity and closed-loop behavioral responsiveness. Every model has a positive success rate drop, yet several strong executors still complete the reference goal frequently under conflict. Moreover, conflict rollouts retain substantial target approach and early reference trajectory structure, with generally limited action magnitude suppression. These signatures remain visible within capability-qualified failures, showing that reference goal noncompletion does not by itself establish behavioral disengagement. Conflict type changes the magnitude and form of the response, while explicit premise checking prompts do not produce a consistent, coordinated shift across outcome and process measurements. We next organize the analysis around four research questions.

\subsection{RQ1: Do Invalid Premises Change Reference Goal Outcomes?}

Table~\ref{tab:overall_conflict_metrics} first separates valid task competence from conflict-conditioned outcomes. PCSR measures success in the reference rollout under consistent premises, while RGSR measures completion of that same reference goal under conflict. All models exhibit a positive $\Delta$SR, ranging from 17.3 to 56.2 percentage points, so invalid premises clearly perturb terminal outcomes. However, the perturbation varies substantially across models. OpenVLA has the largest drop, from a PCSR of 78.7\% to an RGSR of 22.5\%. By contrast, $\pi_{0.5}$, OpenVLA-OFT, and VLA-JEPA combine high PCSR with RGSR values of 77.2\%, 71.1\%, and 69.1\%, respectively. Thus, many conflict rollouts from otherwise capable policies still proceed far enough to complete the reference goal.

These outcome measurements establish whether the reference goal is completed, but not how the policy responds when it is not. A high RGSR is direct evidence of continued reference goal execution. A low RGSR is only evidence of outcome disruption: the policy may depart from the reference behavior, or it may pursue that behavior and fail during perception or control. Consequently, neither $\Delta$SR nor FR is interpreted as a standalone conflict response score.

\subsection{RQ2: How Does Closed-Loop Execution Change under Premise Conflicts?}

The process measurements in Table~\ref{tab:overall_conflict_metrics} compare each conflict rollout with the corresponding reference rollout under consistent premises. Across all matched P0 episodes, TAR remains between 0.765 and 0.896, and Early20 TS remains between 0.784 and 0.886. The result is especially revealing for OpenVLA: although its RGSR falls to 22.5\%, it retains a TAR of 0.765 and an Early20 TS of 0.784. The lower completion rate therefore coexists with substantial observable retention of reference target approach and early execution structure.

For every model, Early20 TS is higher than TS, whose values range from 0.471 to 0.744. This consistent temporal pattern indicates that conflict rollouts initially resemble the reference execution more strongly and diverge later. The measurements cannot identify the internal cause of this divergence, but they are consistent with the policy first initiating familiar task behavior and later accumulating differences during localization, grasping, or control. AMS is also limited, ranging from 0.067 to 0.276. OpenVLA-OFT and $\pi_{0.5}$ have particularly small AMS values of 0.067 and 0.069. Even OpenVLA, which has the largest AMS, retains substantial TAR and Early20 TS. Hence, action magnitude reduction is neither strong nor consistently coordinated with target and trajectory departure.

\begin{table}[t]
\centering
\caption{
Failure rates among capability-qualified pairs and failure-conditioned
mean process measures (\%). Values that are \underline{underlined} are
the most favorable for each metric.
}
\label{tab:failure_conditioned_behavior}

\fontsize{9.0}{9.8}\selectfont
\setlength{\tabcolsep}{2.0pt}
\renewcommand{\arraystretch}{1.08}

\begin{tabular}{cccccc}
\toprule
Model & FR & TAR & Early20 TS & TS & AMS \\
\midrule
OpenVLA & \underline{76.7} & 68.7 & 76.1 & 39.8 & \underline{37.0} \\
OpenVLA-OFT & 28.7 & 64.0 & 69.4 & 33.0 & 19.3 \\
\(\pi_0\) & 41.6 & 68.8 & 74.0 & 34.9 & 13.3 \\
\(\pi_0\)-FAST & 50.2 & 74.1 & 84.5 & 38.7 & 17.8 \\
\(\pi_{0.5}\) & 23.0 & \underline{58.6} & \underline{65.3} & \underline{31.0} & 19.2 \\
GR00T N1.7 & 34.9 & 67.7 & 75.0 & 35.4 & 19.9 \\
VLA-JEPA & 30.5 & 67.4 & 73.4 & 34.3 & 32.0 \\
UniVLA & 34.5 & 70.2 & 80.6 & 40.6 & 20.1 \\
\bottomrule
\end{tabular}

\renewcommand{\arraystretch}{1.0}
\end{table}

\begin{figure}[t]
\centering
\includegraphics[
    width=\columnwidth,
    trim=0mm 0mm 0mm 0mm,
    clip
]{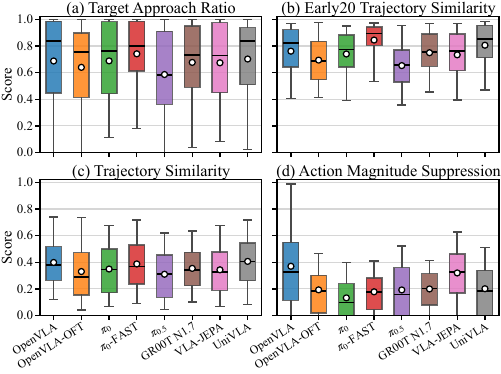}
\caption{
\textbf{Task failure can coexist with substantial retention of reference task behavior.}
Panels (a) to (d) show TAR, Early20 TS, TS, and AMS within capability-qualified failures under P0.
The distributions show that failed conflict rollouts can retain strong target approach and early trajectory similarity.
}
\label{fig:qualified_failure_distributions}
\end{figure}

\subsection{RQ3: Does Reference Goal Failure Imply Behavioral Disengagement?}

To remove episodes in which the model cannot execute the valid task, we analyze capability-qualified failures: the reference rollout under consistent premises succeeds, but the matched conflict rollout does not complete the reference goal. Table~\ref{tab:failure_conditioned_behavior} reports the frequency and behavior of this subset. FR ranges from 23.0\% for $\pi_{0.5}$ to 76.7\% for OpenVLA. Within these failures, however, mean TAR remains between 0.586 and 0.741 and mean Early20 TS remains between 0.653 and 0.845, whereas mean AMS ranges from only 0.133 to 0.370. For example, $\pi_0$-FAST has an FR of 50.2\%, yet its failed rollouts retain a TAR of 0.741 and an Early20 TS of 0.845 with an AMS of 0.178. This is precisely the outcome and process discrepancy hidden by a failure count alone.

Figure~\ref{fig:qualified_failure_distributions} qualifies this conclusion at the episode level. The distributions are broad, so failures are heterogeneous and should not all be labeled as persistence. At the same time, the upper quartile of TAR ranges from 0.898 to 0.985 across models, and the median Early20 TS ranges from 0.657 to 0.894. A substantial subset of failures thus retains strong reference target approach or early trajectory structure. We call this recurring population level pattern \emph{Failed Persistence}: the reference goal is not completed, but observable reference-task behavior remains during execution. This term is a joint outcome and process diagnosis, not a claim that the model understood the conflict, actively refused it, or that every failed episode belongs to a thresholded class.

\begin{figure}[t]
\centering
\includegraphics[width=\columnwidth]{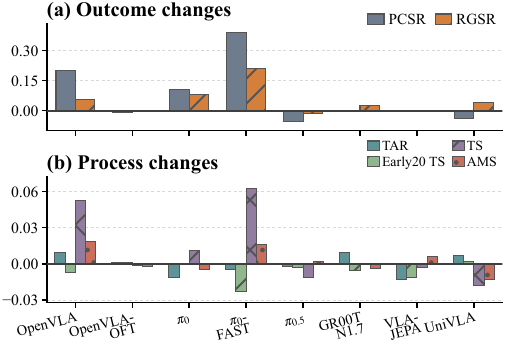}
\caption{
\textbf{Metric differences under explicit premise checking.}
Paired mean differences ($\mathrm{P0}-\mathrm{P1}$) across eight VLAs.
}
\label{fig:p0_p1_changes}
\end{figure}

\subsection{RQ4: How Do Conflict Type and Explicit Premise Checking Affect Behavioral Response?}

Table~\ref{tab:model_by_conflict_taxonomy_main} shows distinct response signatures across the four conflict types. Averaged equally across models, Instruction and Grounding conflicts retain higher RGSR (69.9\% and 64.3\%), TAR (0.938 and 0.923), and TS (0.718 and 0.690) than Spatial and Physical conflicts. Spatial and Physical conflicts produce larger $\Delta$SR values (38.7 and 45.0 percentage points) and lower TAR (0.803 and 0.795), indicating stronger outcome and process disruption on average. The ordering is not universal: $\pi_0$-FAST, for example, has an RGSR of 68.6\% for Spatial conflicts but only 20.8\% for Physical conflicts, whereas several other models exhibit different relative gaps. We therefore treat conflict types as qualitatively different stressors rather than levels of an empirical difficulty scale. Crucially, mean AMS remains low across all four types (0.106--0.127), so greater outcome disruption still does not establish coordinated execution stopping.

Finally, as shown in Figure~\ref{fig:p0_p1_changes}, we compare P0 and P1
to assess whether explicit premise checking produces a selective and
coordinated response. The effects are mixed: only four of eight models
show lower TAR under P1, three show lower Early20 TS, four show lower TS,
and four show higher AMS; no model changes in the intended direction
across all four process measures. For OpenVLA and $\pi_0$-FAST, RGSR
decreases by 5.4 and 20.8 percentage points, while PCSR decreases by
20.2 and 38.8 percentage points, respectively, suggesting general
prompt-induced disruption rather than selective regulation under
conflict. Other models show smaller or opposing changes. Thus, explicit
prompting alone does not consistently produce a coordinated behavioral
response. These paired mean differences are descriptive and are not
presented as tests of statistical significance.

\section{CONCLUSIONS}
\label{sec:conclusion}

In this work, we introduce \textbf{ConflictVLA-Bench}, which uses matched premise-consistent and conflict rollouts to evaluate VLA behavior through outcome and process evidence. All eight representative VLAs show lower original task completion under conflict conditions than under premise-consistent conditions, but they often retain target approach and early trajectory structure with limited action suppression. Even among failed conflict rollouts paired with successful premise-consistent rollouts, mean early trajectory similarity remains above 0.650 for every model, revealing \emph{Failed Persistence} as a recurring behavior across models. Explicit premise checking likewise does not consistently produce coordinated behavioral improvement. Together, these findings show that terminal failure alone does not establish refusal and that strong task execution does not imply reliable responses to invalid premises, underscoring the need to evaluate both outcomes and execution processes.

\addtolength{\textheight}{-12cm}   





\bibliographystyle{IEEEtran} 
\bibliography{main}

@article{kim2024openvla,
  title={Openvla: An open-source vision-language-action model},
  author={Kim, Moo Jin and Pertsch, Karl and Karamcheti, Siddharth and Xiao, Ted and Balakrishna, Ashwin and Nair, Suraj and Rafailov, Rafael and Foster, Ethan and Lam, Grace and Sanketi, Pannag and others},
  journal={arXiv preprint arXiv:2406.09246},
  year={2024}
}

@article{kim2025fine,
  title={Fine-tuning vision-language-action models: Optimizing speed and success},
  author={Kim, Moo Jin and Finn, Chelsea and Liang, Percy},
  journal={arXiv preprint arXiv:2502.19645},
  year={2025}
}

@inproceedings{zitkovich2023rt,
  title={Rt-2: Vision-language-action models transfer web knowledge to robotic control},
  author={Zitkovich, Brianna and Yu, Tianhe and Xu, Sichun and Xu, Peng and Xiao, Ted and Xia, Fei and Wu, Jialin and Wohlhart, Paul and Welker, Stefan and Wahid, Ayzaan and others},
  booktitle={Conference on Robot Learning},
  pages={2165--2183},
  year={2023},
  organization={PMLR}
}

@article{intelligence2025pi_,
  title={{$\pi_{0.5}$}: A Vision-Language-Action Model with Open-World Generalization},
  author={Intelligence, Physical and Black, Kevin and Brown, Noah and Darpinian, James and Dhabalia, Karan and Driess, Danny and Esmail, Adnan and Equi, Michael and Finn, Chelsea and Fusai, Niccolo and others},
  journal={arXiv preprint arXiv:2504.16054},
  year={2025}
}

@article{park2023clara,
  title={Clara: classifying and disambiguating user commands for reliable interactive robotic agents},
  author={Park, Jeongeun and Lim, Seungwon and Lee, Joonhyung and Park, Sangbeom and Chang, Minsuk and Yu, Youngjae and Choi, Sungjoon},
  journal={IEEE Robotics and Automation Letters},
  volume={9},
  number={2},
  pages={1059--1066},
  year={2023},
  publisher={IEEE}
}

@article{black2024pi_0,
  title={{$\pi_0$}: A Vision-Language-Action Flow Model for General Robot Control},
  author={Black, Kevin and Brown, Noah and Driess, Danny and Esmail, Adnan and Equi, Michael and Finn, Chelsea and Fusai, Niccolo and Groom, Lachy and Hausman, Karol and Ichter, Brian and others},
  journal={arXiv preprint arXiv:2410.24164},
  year={2024}
}

@article{bjorck2025gr00t,
  title={Gr00t n1: An open foundation model for generalist humanoid robots},
  author={Bjorck, Johan and Casta{\~n}eda, Fernando and Cherniadev, Nikita and Da, Xingye and Ding, Runyu and Fan, Linxi and Fang, Yu and Fox, Dieter and Hu, Fengyuan and Huang, Spencer and others},
  journal={arXiv preprint arXiv:2503.14734},
  year={2025}
}

@article{james2020rlbench,
  title={Rlbench: The robot learning benchmark \& learning environment},
  author={James, Stephen and Ma, Zicong and Arrojo, David Rovick and Davison, Andrew J},
  journal={IEEE Robotics and Automation Letters},
  volume={5},
  number={2},
  pages={3019--3026},
  year={2020},
  publisher={IEEE}
}

@article{nasiriany2024robocasa,
  title={Robocasa: Large-scale simulation of everyday tasks for generalist robots},
  author={Nasiriany, Soroush and Maddukuri, Abhiram and Zhang, Lance and Parikh, Adeet and Lo, Aaron and Joshi, Abhishek and Mandlekar, Ajay and Zhu, Yuke},
  journal={arXiv preprint arXiv:2406.02523},
  year={2024}
}

@article{liu2023libero,
  title={Libero: Benchmarking knowledge transfer for lifelong robot learning},
  author={Liu, Bo and Zhu, Yifeng and Gao, Chongkai and Feng, Yihao and Liu, Qiang and Zhu, Yuke and Stone, Peter},
  journal={Advances in Neural Information Processing Systems},
  volume={36},
  pages={44776--44791},
  year={2023}
}

@article{chen2026hazardarena,
  title={HazardArena: Evaluating semantic safety in vision-language-action models},
  author={Chen, Zixing and Gao, Yifeng and Wang, Li and Zhao, Yunhan and Liu, Yi and Li, Jiayu and Zheng, Xiang and Wu, Zuxuan and Wang, Cong and Ma, Xingjun and others},
  journal={arXiv preprint arXiv:2604.12447},
  year={2026}
}

@article{fan2026safevla,
  title={SafeVLA-Bench: A Benchmark for the Success-Safety Gap in Vision-Language-Action Models},
  author={Fan, Jialiang and Xu, Weizhe and Sokolsky, Oleg and Lee, Insup and Kong, Fanxin},
  journal={arXiv preprint arXiv:2606.00773},
  year={2026}
}

@article{zhang2026safevla,
  title={Safevla: Towards safety alignment of vision-language-action model via constrained learning},
  author={Zhang, Borong and Zhang, Yuhao and Ji, Jiaming and Lei, Yingshan and Dai, Juntao and Chen, Yuanpei and Yang, Yaodong},
  journal={Advances in Neural Information Processing Systems},
  volume={38},
  pages={153335--153373},
  year={2026}
}

@article{li2025don,
  title={Don’t take the premise for granted: Evaluating the premise critique ability of large language models},
  author={Li, Jinzhe and Li, Gengxu and Chang, Yi and Wu, Yuan},
  journal={arXiv preprint arXiv:2505.23715},
  year={2025}
}

@article{shi2026vla,
  title={VLA-Trace: Diagnosing Vision-Language-Action Models through Representation and Behavior Tracing},
  author={Shi, Haoyuan and Ren, Xiancong and Zhang, Yingji and Zhang, Qinfan and Hu, Jiayu and Shan, Haozhe and Dong, Han and Lu, Jinpeng and Chen, Yinda and Zhang, Yi and others},
  journal={arXiv preprint arXiv:2605.30117},
  year={2026}
}

@article{zhang2026restoring,
  title={Restoring Linguistic Grounding in VLA Models via Train-Free Attention Recalibration},
  author={Zhang, Ninghao and Zhu, Bin and Zhou, Shijie and Chen, Jingjing},
  journal={arXiv preprint arXiv:2603.06001},
  year={2026}
}

@article{bu2025univla,
  title={Univla: Learning to act anywhere with task-centric latent actions},
  author={Bu, Qingwen and Yang, Yanting and Cai, Jisong and Gao, Shenyuan and Ren, Guanghui and Yao, Maoqing and Luo, Ping and Li, Hongyang},
  journal={arXiv preprint arXiv:2505.06111},
  year={2025}
}

@article{sun2026vla,
  title={Vla-jepa: Enhancing vision-language-action model with latent world model},
  author={Sun, Jingwen and Zhang, Wenyao and Qi, Zekun and Ren, Shaojie and Liu, Zezhi and Zhu, Hanxin and Sun, Guangzhong and Jin, Xin and Chen, Zhibo},
  journal={arXiv preprint arXiv:2602.10098},
  year={2026}
}

@article{pertsch2025fast,
  title={Fast: Efficient action tokenization for vision-language-action models},
  author={Pertsch, Karl and Stachowicz, Kyle and Ichter, Brian and Driess, Danny and Nair, Suraj and Vuong, Quan and Mees, Oier and Finn, Chelsea and Levine, Sergey},
  journal={arXiv preprint arXiv:2501.09747},
  year={2025}
}

@misc{zhang2025vlaarenaopensourceframeworkbenchmarking,
      title={VLA-Arena: An Open-Source Framework for Benchmarking Vision-Language-Action Models}, 
      author={Borong Zhang and Jiahao Li and Jiachen Shen and Yishuai Cai and Yuhao Zhang and Yuanpei Chen and Juntao Dai and Jiaming Ji and Yaodong Yang},
      year={2025},
      eprint={2512.22539},
      archivePrefix={arXiv},
      primaryClass={cs.RO},
      url={https://arxiv.org/abs/2512.22539}, 
}

@article{mees2022calvin,
  title={Calvin: A benchmark for language-conditioned policy learning for long-horizon robot manipulation tasks},
  author={Mees, Oier and Hermann, Lukas and Rosete-Beas, Erick and Burgard, Wolfram},
  journal={IEEE Robotics and Automation Letters},
  volume={7},
  number={3},
  pages={7327--7334},
  year={2022},
  publisher={IEEE}
}

@article{ahn2022can,
  title={Do as i can, not as i say: Grounding language in robotic affordances},
  author={Ahn, Michael and Brohan, Anthony and Brown, Noah and Chebotar, Yevgen and Cortes, Omar and David, Byron and Finn, Chelsea and Fu, Chuyuan and Gopalakrishnan, Keerthana and Hausman, Karol and others},
  journal={arXiv preprint arXiv:2204.01691},
  year={2022}
}

@article{ren2023robots,
  title={Robots that ask for help: Uncertainty alignment for large language model planners},
  author={Ren, Allen Z and Dixit, Anushri and Bodrova, Alexandra and Singh, Sumeet and Tu, Stephen and Brown, Noah and Xu, Peng and Takayama, Leila and Xia, Fei and Varley, Jake and others},
  journal={arXiv preprint arXiv:2307.01928},
  year={2023}
}

@article{huang2022inner,
  title={Inner monologue: Embodied reasoning through planning with language models},
  author={Huang, Wenlong and Xia, Fei and Xiao, Ted and Chan, Harris and Liang, Jacky and Florence, Pete and Zeng, Andy and Tompson, Jonathan and Mordatch, Igor and Chebotar, Yevgen and others},
  journal={arXiv preprint arXiv:2207.05608},
  year={2022}
}

@misc{zhou2025liberoprorobustfairevaluation,
      title={LIBERO-PRO: Towards Robust and Fair Evaluation of Vision-Language-Action Models Beyond Memorization}, 
      author={Xueyang Zhou and Yangming Xu and Guiyao Tie and Yongchao Chen and Guowen Zhang and Duanfeng Chu and Pan Zhou and Lichao Sun},
      year={2025},
      eprint={2510.03827},
      archivePrefix={arXiv},
      primaryClass={cs.CV},
      url={https://arxiv.org/abs/2510.03827}, 
}

@misc{fei2025liberoplusindepthrobustnessanalysis,
      title={LIBERO-Plus: In-depth Robustness Analysis of Vision-Language-Action Models}, 
      author={Senyu Fei and Siyin Wang and Junhao Shi and Zihao Dai and Jikun Cai and Pengfang Qian and Li Ji and Xinzhe He and Shiduo Zhang and Zhaoye Fei and Jinlan Fu and Jingjing Gong and Xipeng Qiu},
      year={2025},
      eprint={2510.13626},
      archivePrefix={arXiv},
      primaryClass={cs.RO},
      url={https://arxiv.org/abs/2510.13626}, 
}

@article{fang2026vision,
  title={When vision overrides language: Evaluating and mitigating counterfactual failures in vlas},
  author={Fang, Yu and Feng, Yuchun and Jing, Dong and Liu, Jiaqi and Yang, Yue and Wei, Zhenyu and Szafir, Daniel and Ding, Mingyu},
  journal={arXiv preprint arXiv:2602.17659},
  year={2026}
}

@inproceedings{hsieh2025teaching,
  title={Do What? Teaching Vision-Language-Action Models to Reject the Impossible.},
  author={Hsieh, Wen-Han and Hsieh, Elvis and Niu, Dantong and Darrell, Trevor and Herzig, Roei and Chan, David M},
  booktitle={EMNLP (Findings)},
  pages={11861--11869},
  year={2025}
}

@article{yeke2026yes,
  title={The Yes-Man Syndrome: Benchmarking Abstention in Embodied Robotic Agents},
  author={Yeke, Doguhan and Temirel, Elif Su and Shreekumar, Ananth and Lee, Brandon and Xu, Dongyan and Celik, Z Berkay},
  journal={arXiv preprint arXiv:2605.20544},
  year={2026}
}

\end{document}